\documentclass[nonacm]{acmart}

\def\BibTeX{{\rm B\kern-.05em{\sc i\kern-.025em b}\kern-.08emT\kern-.1667em\lower.7ex\hbox{E}\kern-.125emX}}
    
\copyrightyear{2019}
\acmYear{2019}
\setcopyright{acmlicensed}
\acmConference[HT '19]{HT '19: 30th conference on Hypertext and Social Media}{September 17--20, 2019}{Hof, Germany}
\acmBooktitle{HT '19: Proceedings of the 30th on Hypertext and Social Media, September 17--20, 2019, Hof, Germany}
\acmPrice{XX.XX}
\acmDOI{10.XXXX/XXXXXXX.XXXXXXX}
\acmISBN{XXX-X-XXXX-XXXX-X/XX/XX}

\usepackage{tipa}

\begin{document}

\fancyhead{}

%
\title[Multi-Modal Citizen Science]{Multi-Modal Citizen Science: From Disambiguation to Transcription of Classical Literature}

%

\author{Maryam Foradi}
\authornotemark[1]
\email{maryam.foradi@uni-leipzig.de
}
\affiliation{%
  \institution{Leipzig University}
  \streetaddress{Augustusplatz 10}
  \city{Leipzig}
  \state{Germany}
  \postcode{04109}
}

\author{Jan Ka\ss{}el}
\authornote{All four authors contributed equally to this research.}
\email{jan.kassel@studserv.uni-leipzig.de}
\affiliation{%
  \institution{Leipzig University}
  \streetaddress{Augustusplatz 10}
  \city{Leipzig}
  \state{Germany}
  \postcode{04109}
}

\author{Johannes Pein}
\authornotemark[1]
\email{johannes.pein@studserv.uni-leipzig.de
}
\affiliation{%
  \institution{Leipzig University}
  \streetaddress{Augustusplatz 10}
  \city{Leipzig}
  \state{Germany}
  \postcode{04109}
}

\author{Gregory R. Crane}
\authornotemark[1]
\email{gregory.crane@tufts.edu
}
\affiliation{%
  \institution{Tufts University}
  \city{Medford}
  \state{USA}
  \postcode{MA 021554109}
}

%
\renewcommand{\shortauthors}{Foradi et al.}

%
\begin{abstract}
The engagement of citizens in the research projects, including Digital Humanities projects, has risen in prominence in the recent years. This type of engagement not only leads to incidental learning of the
participant, but also indicates the added value of corpus enrichment via different types of annotations undertaken by users generating the so-called smart texts. Our work focuses on the continuous task
of adding new layers of annotation to Classical Literature from the around the world. We aim to provide more extensive tools for readers of smart texts, enhancing their reading comprehension and at the same time empowering the language learning by introducing intellectual tasks, i.e. linking, tagging, and disambiguation. The current study adds a new mode of annotation, audio annotations,
to the extensively annotated corpus of poetry by the Persian poet Hafiz, proposing tasks with three different difficulty levels to estimate users' ability in order to rate their annotation in further stages
of the project, where no ground truth is available. Annotators with no knowledge of Persian are able to add annotations to the Persian source.

\end{abstract}

%
%
\ccsdesc[500]{Information systems~Crowdsourcing}
\ccsdesc[500]{Applied computing~Interactive learning environments}
\ccsdesc[300]{Human-centered computing~User studies}
%
\keywords{smart texts; citizen science; user experience design; computer assisted language learning (CALL)}
%
\maketitle
\section{Introduction}


The increasingly growing adoption of Citizen Science (CS) in academic projects by not only the researchers from STEM disciplines but also (digital) humanists acceleratingly attracts non-professionals from all ages and skills. In the humanities this contribution, with tasks such as text transcription, corrections, classifications, tagging and other annotations, leads to an enrichment of digitized or digitally-born corpora with different types of annotations. However, in most of the CS projects education is not considered a high-priority goal, but rather a side-effect that happens out of the context of the project \cite{veja2018, lee2016}. On the opposite hand, projects labeled as education projects aim at achieving educational results while the scientific output is considered as a secondary goal \cite{Wiggins:2011}.

Although crowdsourcing in Natural Language Processing (NLP) profits from the engagement of the members of public as cheap labourers finishing micro-tasks, the cognitive engagement of them is mostly underestimated. The motivation for participation in such projects is often limited to the micro-payments offered via scholars on platforms such as Amazon Mechanical Turk\footnote{\url{https://www.mturk.com/}}, which can lead to a demographic bias effecting the accuracy of the data \cite{ross2010}. Yet, the recently introduced concept of learnersourcing \cite{kim2015} is an endeavour to provide an opportunity for annotating the learning materials by learners for the other learners and increasing the learning potential of learnersourcers.

In this paper we propose a novel research idea alongside a prototype application\footnote{\url{https://git.informatik.uni-leipzig.de/dsr/hafiz-prototype}} to investigate the extent to which learnersourcing facilitates the phonetic transcription of Persian corpora so that not only the learnersourcers can practice their listening skills, but also other learners have access to more accurate phonetic transcriptions by the means of this medium.

One of the most challenging issues in the digital processing of Arabic and Persian corpora are short vowels (/\ae/, /e/ and /o/) that are spoken but not explicitly written. This results in various words that have the same written appearance, but possess entirely different pronunciations, meanings or grammatical features \cite{shamsfard2011challenges}. In this context, the problem is not limited to the automatic phonetic transcription, in which the absence of short vowels in the replacement of Arabic characters with Latin characters leads to the poor readability of transliterated text. Even the existence of a data-set including words written in original language linked with their manual phonetic transcription is deceptive, because the correct form of transcription cannot be identified unless within the context.

While methods of machine transliteration between Arabic and English exist \cite{karimi_machine_2008},  we propose a different, learnersourced approach to this problem. By means of already existing audio recordings of classical texts linked to the source text, we aim at a) choosing the correct form of phonetic transcription among the existing manually transliterated options, and b) adding the short vowels to the automatic transcriptions, where the short vowels are missing. Although the recordings shall be produced by experts or native speakers of the named languages, the phonetic transcription task can be performed by both language learners and interested readers, who are unfamiliar with the languages as such.

In light of the above, we propose an application that allows the users to improve their understanding of the Persian poetry and language by engaging them with interactive tasks while also gathering data correlating the corrections to our transliterated corpus with the linguistic proficiency of our participants.  Albeit the evaluation of the pedagogic influence of our application exceeds the limitations of the current effort, we pose the following questions:

\vspace{0.5em}

\noindent 1) Can accurate corrections of a phonetically transcribed corpus be acquired through our application in the form of learnersourcing?

\vspace{0.15em}

\noindent 2) How can the complexity of specific task be measured to allow for a more structured learnersourcing experience? 

\vspace{0.15em}

\noindent 3) What are the pedagogical benefits of this type of activity, in terms of improved vocabulary learning with the audio annotations?

\section{Related Work}

The problem of text-to-speech alignment (phonetic alignment) is approached with several methods such as use of Hidden Markov Models for alignment of speech on text \cite{talkin1994,Coile1994} or the combination of text-to-speech systems and the Dynamic-Wrapping algorithm \cite{Malfrre1997}. These attempts are mostly limited to the languages such as English, French, German, and Spanish. However, there are efforts to expand this field by adding more languages \cite{Goldman2010}. The rise of crowdsourcing provided the opportunity for researchers to create multilingual transcribed corpora linked to the audio annotations \cite{Callison-Burch:2010,Caines2016}, but also to include low-resource languages, to avoid the slow and cost intensive process of transcriptions by experts \cite{gelas2011}. Although the studies proved that the crowdworkers can achieve high data accuracy \cite{williams2011,gelas2011}, to our knowledge the task is mostly running on platforms such as Amazon Mechanical Turk, where micro-payments are offered to workers for performing micro-tasks.

The newly introduced concept of learnersourcing \cite{kim2015} describes the use of input by larger groups of learners to improve the learning video contents and interfaces. Some related projects in this context that also pursue the same objective aim at creating learnersourced subgoals in the framework of how-to videos \cite{Weir:2015:LSL:2675133.2675219}, while some others crowdsource the assessments of exercises \cite{Mitros.2015} or provide personalized hints \cite{Glassmann.2016} and generate explanations \cite{Williams.2016} for learners supporting them to solve the problems. Furthermore, in the field of semantic annotations studies are conducted to understand the effectiveness of learnersourcing through the inter-contextual linking of Quranic texts \cite{Basharat:2016}. Though, the most akin project to our study is CrowdClass by \citeauthor{lee2016}, which allows the same concept of on-task learning as intended in our project \cite{lee2016}. CrowdClass integrates the learning of scientific concepts through the so-called scaffold learning with the task of image classification. In this study the performance of the citizen scientist is investigated and the potential effect of learning while working on classification of astronomical images similar to the one of GalaxyZoo \footnote{\url{http://zoo1.galaxyzoo.org/}} is evaluated.

Studies in the field of language learning show that the different types of annotation leverage the language acquisition \cite{Abraham2008}, \cite{Claros2009}, and audio annotations are beneficial for vocabulary learning due to the fact that in this way the sound form is captured by the phonological memory \cite{hummel2010}. The audio annotations play a meaningful role specially for the learners who are not familiar with the alphabets and phonemes of the language they learn \cite{dubios2001}, \cite{laufer2000}. But yet, to our knowledge there is no project that attempts to disambiguate or correct phonetic transcriptions by non-native speakers and language learners enabling them not only to learn the language with the help of audio annotations but also provide support materials that can be useful for other learners.


\section{Research Method}

\subsection{Research Design}
\label{sec/design}

\begin{figure}[h]
  \centering
  \includegraphics[width=.6\linewidth]{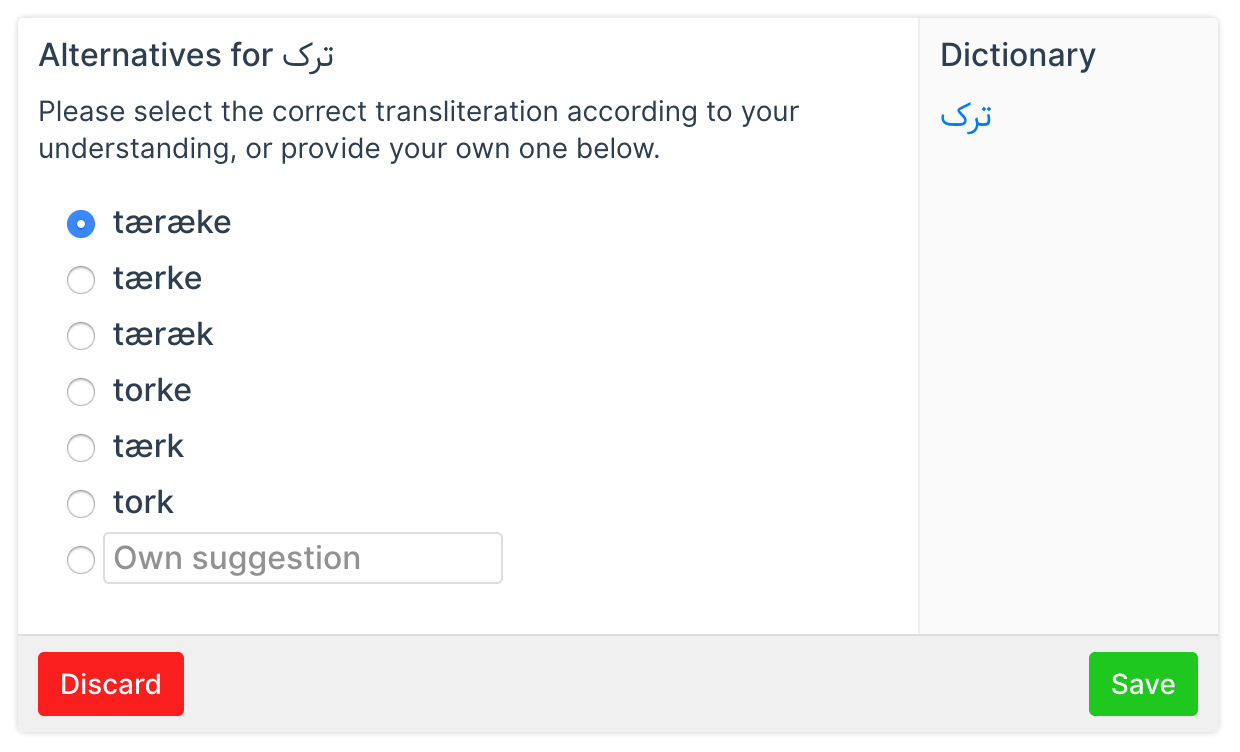}
  \caption{User interface for choosing the correct phonetic transcription according to the audio.}
  \Description{User interface for the transcription task.}
  \label{fig/ui}
\end{figure}

In this project, we investigate the extent to which the users with no knowledge of Persian are able to correct the phonetic transcriptions while listening to audio recordings. For this purpose, we confront the participant with disambiguation/correction tasks of varying complexity. In each task, the participant is provided with a range of plausible IPA-based transcriptions and an input field to add a complementary, probably more fitting phonetic transcription. These tasks can be grouped into a) disambiguation tasks, where the correct transcription is given as an option, and b) completion tasks, in which the users are supposed to add only the short vowels and c) correction tasks, where the correct transcription is missing, and it ought to be detected and inserted manually by the users. From technical point of view, the complexity of tasks is classified, depending on a) whether a ground truth, provided by a language expert exists b) whether a data-set with links between the original text and the possible phonetic transcriptions is available and c) whether all options are machine generated, while all the short vowels are missing and the correct phonetic transcription is unknown.

While the first type of task shall be used to evaluate the language capabilities and listening skill of the user, the second and third types represent the actual and proposed learnersourcing by leaving the finding of the correct phonetic form to the crowd of learners. Before working on these tasks, the users are presented with an introductory text explaining the participation process and the basic concept of the tasks. The complexity and classification of the specific tasks as described above is unknown to the user. To answer the first research question addressed above we conducted a pilot study that is described in detail in below sections.

\subsection{Data Collection}
For our pilot study, we designed an interface designed and provided the users with six lines of Persian poetry with the original text written in Arabic letters, the corresponding IPA-based phonetic transcription \cite{ipa1999} of the text, and the audio recording by one native speaker reading the text. For each recording, a speech-to-text service is used to create a word-by-word alignment of the source text and audio recording through time-stamps. Depending on whether there are different options for the pronunciation of the words, we inserted intentionally possible but wrong pronunciation for some random words in the displayed transliterated text. The users were confronted with 20 disambiguation items (in form of disambiguation by clicking on the correct option among the multiple choice options) and 2 correction items (by typing the correct form in the corresponding field). For all 22 items, a ground truth was available.

To allow a deeper analysis of the data gathered, the users were asked to create a profile before solving the tasks. In the profile, the user's L1 and L2 languages could be entered as well as the information about the user's age, gender, education and nationality. Afterwards, we asked the users to listen to the audio recordings and for those words, which were part of the exercise, to choose the correct pronunciation by clicking on the correct answer in the multiple choice pop-up window, as shown in Figure~\ref{fig/ui}. While users listened to audio, the corresponding word and its phonetic transcription turned into red, and they were able to pause, repeat or start over the audio play. 
By solving the phonetic transcription task in the form of disambiguation or manual correction, the users generated a database with possible corrections that can be used for various purposes, such as the enhancement of machine transliteration accuracy as discussed in the context of transliterations between Persian and English in \cite{karimi_machine_2008}.





\subsection{Modeling of Phonetic Distances}

As outlined in ~\ref{sec/design}, we provided participants with various options of IPA-based transcriptions to choose from, so that we can calculate the error rate for each item. A plain binary correctness grading would certainly underestimate the complexity of each task, as the difficulty and complexity of the task may depend on the distinctive features of the to-be-changed phones. For instance, two consonants /t/ and /s/ share more phonological features than /m/ and /\textdyoghlig/. In the first pair there are only two differences in Manner features, but the latter pair has differences in all Class, Place, Laryngeal and Manner features \cite{katamba1989}, which means that distinguishing between /t/ and /s/ shall be more challenging than between /m/ and /\textdyoghlig/. This demands the calculation of distance weightings, which enables us to explore the correlation between the average error rate for each item and the estimated correction effort, i.e. the distance between the displayed form and the correct answer, for that particular item.

\citeauthor{fontan2016} provide a multifaceted approach of calculating distances between phonological features, by calculating phonologically weighted Levenshtein's distance (PWLD) \cite{fontan2016}. Based on a mapping of phonetic features of the IPA letters by \citeauthor{katamba1989} \cite{katamba1989}, we calculate distances for all phonetic transcription options for consonants and vowels, respectively. Since not all the spoken vowels are existence in Persian phonology \cite{miller2012}, feature mappings for vowels has been adapted to Persian, as just a subset of the features described by \citeauthor{katamba1989} is used in Persian language.

For calculating the distance measure~$d$, we proceeded as follows: Having a ground truth phonetic transcription~$w$ and a given answer~$u$, for each letter, the basic distance is equal to the number of different phonetic features~$p$ divided by the total number of phonetic features~$n$ \cite{fontan2016}:

\begin{equation}
    d(w, u) = \frac{p(w, u)}{n}
\end{equation}

The total number of phonetic features for consonants and vowels is 15 and 4, respectively. However, a stark weighting of a single feature difference occurs when treating vowels, which would result in a distance of $\frac{1}{4} = 0.25$ due to just 4 phonetic features related to vowels, as opposed to the 15 features related to consonants, $\frac{1}{15} \approx 0.067$. Thus, we employed a simple quadratic easing for vowel distances, $d_v$:

\begin{equation}
  d_v(w, u) = d^2(w, u)
\end{equation}

This results in $(\frac{1}{4})^2 = 0.0625$ and increases squarely as more vowel-related features differ.

\subsection{Technical Implementation}
\label{sec/implementation}

When approaching the realization of the research method, we were facing three technical tasks: First, the corpus of Persian poetry we conducted was lacking audio annotation alignments. Second, this corpus was to be presented to users via a user interface during user studies. And third, collected user data and study results were to be stored and, subsequently, evaluated within a database.

To gather audio annotation alignment data, we utilized the Google Cloud Text-to-Speech\footnote{\url{https://cloud.google.com/speech-to-text/}} service. By using this service with our prerecorded spoken audio, configured to recognize Persian language, word-level alignments were acquired quickly for all of our corpus within short time. As the data lacked some inaccuracies, however, manual adjustments were necessary.

As our initial motivation arose from the Perseus Project \cite{smith2000}, and, especially, its Scaife Viewer\footnote{\url{https://scaife.perseus.org/}} application, we opted to build a JavaScript-based single-page application (SPA) for users to interact with. The user interface library Vue.js\footnote{\url{https://vuejs.org/}} has been adopted for this work, as it promised possible interoperability with Scaife Viewer for incorporating the audio alignment functionality. Data storage is provided by an Apache CouchDB\footnote{\url{https://couchdb.apache.org/}} NoSQL database, which offers convenient interaction capabilities via its HTTP interface. The data analysis pipeline then converts the filtered and enriched data into the CSV-based tabular data for further post-processing and analysis.

\section{Pilot Study Results}

As stated in previous sections, due to the novelty of this project idea we conducted a pilot study to evaluate the methodology of data collection and refine the research questions. Hence, in scaffolding learning the adjustment of the scaffolding is one of the most essential and challenging steps. The prototype application went live for one week, during which a total number of 31 volunteers participated in the experiment. Of these 31 submissions only 16 are valid: Some submissions either contained corrupt data, or participants stated their knowledge of the Persian language, which meant that they did not fit within the scope of the study. Among the remaining 16 participants, four provided us with no profile, five of them entered German as their native language, four English, two French, and one Russian. Finally, none of these considered participants stated any familiarity with Persian.

For the analysis we identified three independent variables for each item; two of them were interval variables namely word length and the PWLD between the displayed text and the correct answer, i.e. what the user hears in the audio. The other parameter was a nominal parameter, which revealed if the correct answer was among the to-be-selected options or not. We weighted each item with the number of times that an answer was given to the option, and we explored the effect of the named parameters on the weighted error rate for each item.

A multiple linear regression was calculated to predict error rate based on PWLD, word length and the existence of correct answers among the options. A significant regression equation was found ($F(3,216) = 135.238, p < 0.001$), with an $R^2$ of $0.653$. Items' predicted error rate is equal to ${-0.425} + 3.918$ (PWLD) ${-0.142}$ (word length) $ +2.545$ (existence of the correct answer), where distance is measured as outlined above, length increased with each added character and the existence of correct answer is coded as $1 =$ existence, $2 =$ non-existence. Item's error rate increased for 3.918 depending on the increase of distance. Error rate decreased $.142$ for one character more in the length of the word, and it was $2.545$ higher, when the correct answer was not among the presented options. PWLD and the existence of the correct answer were significant predictors of error rate, whereas the word length was not a significant factor. 

\section{Discussion}


While the application created to conduct this study is still in the early stages of development, both the functionality and the user interface have been improved based on the participants' feedback. As the data analyzed in this study has been gathered throughout the development process, these inconsistencies in the workflow of our application are also adversely affecting the reliability of our data. Furthermore, the presented data is the result of a pilot study with one group design. The objective of this pre-experimental study was finding the deficiencies in the research design which helps us to improve the design and adjust our scaffolding. As the results of the pilot study indicate, the highest accuracy is achieved, when the displayed text is the correct answer, but when the displayed text is not the correct option, and it has to be selected among the options of multiple choice test, the accuracy drops down to 57.7\%. However, if we decrease the complexity of the items to the lowest possible, i.e. eliminating the items with more than two options, and also the items, where the correct answer shall be entered manually, we achieve an accuracy equal to 67.8\%. Alone this, emphasizes the importance of the scaffolding, which needs to be improved in the further stages.

Our initial research design also, as it happened, limited the capacity of our data. While some of the results were gathered with staff familiar with the project supporting the users in person, some users took part online without any assistance beyond the introductory text. From the feedback gathered, the in-person support was specially beneficial to users unfamiliar with the Persian language and the IPA. Additionally, creating a profile was optional, as was the amount of tasks the users solved before submitting the results. This also negatively influenced the reliability of our data. In future research, both a profile with a minimal set of information and completing all tasks presented to the user will be mandatory.
If we want to further allow online participation, we will have to indicate these contributions to distinct them from results of on-site participation.
Adding an interactive tutorial introducing the workflow of our application and the IPA to the user will also help to reduce the need for in-person support and allow for an preliminary evaluation of the users relevant capabilities.

\section{Conclusions \& Future Work}

While our application provides the users with interactive tasks that we presume to have an positive effect on their learning experience, we need to develop the pedagogical structure to further the beneficial impact on their language capabilities. In order to grade the user's results, we use a modified algorithm for phonologically weighted Levenshtein's distance, which assigns a complexity to each option of a tasks based on it's ground truth. The results of the pilot study indicate that this distance has a significant effect on the error rate of the items, however, the extent of this impact and further parameters are still unknown to us. Conclusively, by evaluating these parameters we aim at creating a rating system, allowing us to order the tasks by complexity and present users with tasks matching their actual language capabilities, thus helping to create a more structured learnersourcing experience. Alleviated by this transformation, further research on the operationality of learnersourcing for the procuration of corrections to phonetically transliterated corpora will be facilitated.

As initially stated, the evaluation of the effect of our application on the understanding of Persian language, poetry and culture exceeded the limitations of this pre-experimental study. To permit the classification of the application as learnersourcing, additional research needs to be done in the form of two experiments. One experiment focuses on the influence of this kind of problem solving on the language capabilities of the users as a possible starting point. By analyzing the differences between vocabulary learning through this learnersourcing application with a `common' vocabulary learning technique like flashcards with audio annotations, we can discover the prospects and measures necessary to further transform the learning experience of the users to their benefit. In this way, we will obtain a control group for comparison, which leads to the increase of reliability and validity of the study.

This experiment shall also be combined with a second experiment, in which the performance of `regular' language learners on solving the defined task will be compared with the performance of the users with no background knowledge in that particular language in terms of data accuracy. The experiment still requires an appropriate scaffolding, consisting of the already mentioned tutorial in combination with protocol for the in person support provided during  participation. In this regard, we believe that familiarity of users with the IPA and also adding some out-of-test exercises (e.g., a tutorial) will enable the users to perform more accurately during the testing phase.

The application itself is at an early stage of development. While all tasks used in this study were created manually by an expert, the intended learnersourcing of corrections to a transliterated corpus requires the machine generation of multiple, plausible, distinct transcription based on phonetic transcriptions of the users. While we have the ground truth for the initial evaluation of the users available, a rating system will be integrated into our application, so that the performance of the users can be voted where the task is not limited to disambiguation anymore and the transcription is required and no ground truth is available.

\begin{acks}
We would like to thank all the individuals who shared their time for participating in the preliminary study described in this article. 
\end{acks}

%
\bibliographystyle{ACM-Reference-Format}
\bibliography{references}

%




\end{document}